\documentclass[9pt]{article}

\usepackage{spconf,amsmath,graphicx}

\usepackage{cite}
\usepackage{amssymb,amsfonts}
\usepackage{algorithmic}
\usepackage{textcomp}
\usepackage{xcolor}
\usepackage{stfloats}
\usepackage{algorithm}

\usepackage{booktabs}  
\usepackage{threeparttable}
\usepackage{multirow}

\usepackage{epsfig}
\usepackage{threeparttable}

\usepackage{amssymb}
\usepackage{verbatim}
\usepackage{subfigure}
\usepackage{graphicx}
\usepackage{float}
\usepackage{stfloats}
\usepackage{flushend}

\DeclareMathOperator*{\argmax}{argmax}

\title{Adaptive Adjustment with Semantic Feature Space\\
for Zero-Shot Recognition}
\name{Jingcai Guo, Song Guo\thanks{This work is supported by National Natural Science Foundation of China (61872310) and INTPART BDEM project.}}
\address{Department of Computing, The Hong Kong Polytechnic University, Hong Kong\\
\texttt{cscjguo@comp.polyu.edu.hk, song.guo@polyu.edu.hk}}
\begin{document}
%
\maketitle
\begin{abstract}
In most recent years, zero-shot recognition (ZSR) has gained increasing attention in machine learning and image processing fields. It aims at recognizing unseen class instances with knowledge transferred from seen classes. This is typically achieved by exploiting a pre-defined semantic feature space (FS), i.e., semantic attributes or word vectors, as a bridge to transfer knowledge between seen and unseen classes. However, due to the absence of unseen classes during training, the conventional ZSR easily suffers from domain shift and hubness problems. In this paper, we propose a novel ZSR learning framework that can handle these two issues well by adaptively adjusting semantic FS. To the best of our knowledge, our work is the first to consider the adaptive adjustment of semantic FS in ZSR. Moreover, our solution can be formulated to a more efficient framework that significantly boosts the training. Extensive experiments show the remarkable performance improvement of our model compared with other existing methods.
\end{abstract}
\section{Introduction and Related Work}
Zero-shot recognition (ZSR) imitates human ability in recognizing new unseen classes. It is achieved by exploiting labeled seen class instances and certain knowledge that is shared between seen and unseen classes \cite{lampert2014attribute,rahman2018unified}. This knowledge, i.e., attributes, exists in a high dimensional vector space called semantic feature space (FS). The attributes are meaningful high-level information about instances such as their shapes, colors, components, textures, etc. Semantic features describe a class or an instance, in contrast to the typical classification, which names an instance. Intuitively, the similar classes have similar patterns in the semantic FS. These particular patterns are called prototypes. In ZSR, the common practice is first to map an unseen class instance from its original FS, i.e., visual FS, to semantic FS by a mapping function trained on seen classes. Then with such semantic features, we search its most closely related prototype whose corresponding class is set to this instance. 

\begin{figure}[tp]
	\centerline{\includegraphics[width=2.9in]{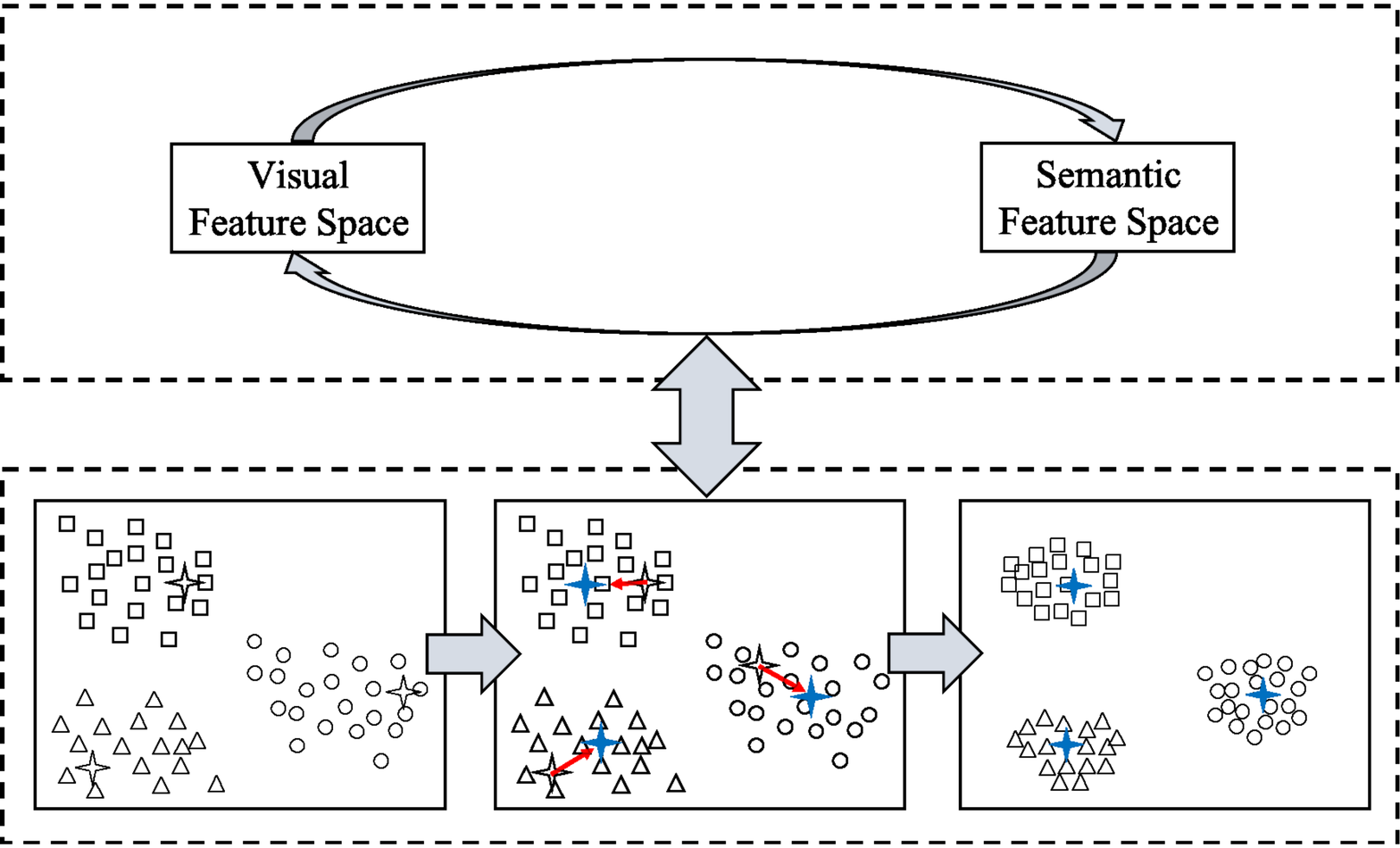}}
	\caption{Unified Framework}
	\label{fig1}
\end{figure}

However, as one of the key building blocks in ZSR, the mapping function is trained solely on seen classes. Although the knowledge is shared by both seen and unseen classes, the training and testing classes are intuitively different. Due to the absence of unseen classes during training, ZSR easily suffers from the domain shift problem \cite{fu2015transductive} which refers to the phenomenon that when mapping unseen class instances from their visual to semantic FS, the obtained results may shift away from the real ones (prototypes). Moreover, during searching step, a small number of prototypes may easily become the most related prototypes to most testing unseen class instances. This challenge is the so-called hubness problem \cite{shigeto2015ridge}. 

To deal with these issues, several transductive learning based methods \cite{fu2015transductive} assume that the unseen class instances (unlabelled) are available at once during training. DeViSE \cite{frome2013devise} trains a linear mapping between visual and semantic FS by an effective ranking loss formulation. ESZSL \cite{romera2015embarrassingly} utilizes the square loss to learn the bilinear compatibility and adds regularization to the objective with respect to Frobenius norm. SSE \cite{zhang2015zero} uses the mixture of seen class parts as the intermediate FS. AMP \cite{bucher2016improving} embeds the visual features into the attribute space. SynC$^{struct}$ \cite{changpinyo2016synthesized} and CLN+KRR \cite{long2017zero} jointly embed several kinds of textual features and visual features to ground attributes. MFMR \cite{xu2017matrix} leverages the sophisticated technique of matrix tri-factorization with manifold regularizers to enhance the mapping between visual and semantic FS. With the popularity of generative adversarial networks (GANs), GANZrl \cite{tong2018adversarial} applies GANs to synthesize instances with specified semantics to cover a higher diversity of seen classes. Instead, GAZSL \cite{zhu2018generative} leverages GANs to imagine unseen classes from text descriptions. Despite the efforts made, the domain shift and hubness problems are still open issues.

In this paper, we propose a novel model based on a specific learning framework to adaptively adjust the semantic FS by considering both the prototypes and the global distribution of data (Fig.1, bottom). Specifically, some key building blocks of ZSR, i.e., the semantic FS and the distribution of data, do not seem to receive comparable attention. Conventional ZSR models normally regard the pre-defined semantic FS as unchangeable and keep each prototype fixed during training. However, we observe that when mapping unseen class instances to semantic FS, the obtained features are quite concentrated. Furthermore, some pre-defined prototypes are also too closely distributed. These deficiencies affect models' ability to adapt and generalize to unseen classes. As we know, the process of human beings understanding things is constantly improving. Similarly, we argue the semantic FS also needs to be adjusted in order to mitigate the domain shift and hubness problems. Moreover, we propose to combine the adjustment with a cycle mapping, which first maps the instance from visual to semantic FS and then vice versa, guaranteeing that the mapping obtains more robust results to further alleviate the domain shift problem (Fig.1, top). We formulate the above steps to a more efficient framework that significantly boosts the training of ZSR. 

\section{Proposed Approach}
\subsection{Cycle Mapping}
In zero-shot recognition, for the visual to semantic mapping, the recognition can be described as:
\begin{small}
\begin{equation}
\begin{aligned}
c\left ( x^{(u)}_{i} \right ) = \mathop{\argmax}_{c\epsilon C^{(u)}} \Omega \left ( f_{v\rightarrow s}\left ( x^{(u)}_{i} \right ), p^{(u)} \right ),
\end{aligned}
\end{equation}
\end{small} 

\noindent where $c\left ( x^{(u)}_{i} \right )$ is the predicted class of unseen class instance $x^{(u)}_{i}$ that belongs to unseen classe set $C^{(u)}$, $\Omega(\cdot,\cdot )$ is a similarity measurement, $p^{(u)}$ is the unseen class prototypes and $f_{v\rightarrow s}(\cdot)$ is the mapping function trained on labeled seen classes that maps from visual to semantic feature space. The training can be described as:
\begin{small}
\begin{equation}
\min \sum_{i=1}^{m} \left \| f_{v\rightarrow s}\left ( x^{(s)}_{i} \right ) - p_i^{(s)} \right \|^{2},
\end{equation}
\end{small}

\noindent where $m$ is the number of seen class instances, $x^{s}_{i}$ is the $i$-th instance of seen classes and $p_i^{(s)}$ is the prototype corresponding to $x^{s}_{i}$. Similarly, for the visual to semantic mapping, we aim to find a mapping that reversely maps the semantic to visual FS. In our model, we combine both mapping directions to a cycle mapping. It can be formulated to an encoder-decoder structure $f_{v \rightleftharpoons s}\left ( x^{(s)} \right ) = f_{s\rightarrow v}\left ( f_{v\rightarrow s}\left ( x^{(s)} \right ) \right )$. The training can be described as:
\begin{small}
\begin{equation}
\begin{split}
\min \sum_{i=1}^{m} \left \| x^{(s)}_{i} - f_{v \rightleftharpoons s}\left ( x^{(s)}_{i} \right ) \right \|^{2}, s.t.\quad f_{v\rightarrow s}\left ( x^{(s)}_{i} \right ) = p_i^{(s)},
\end{split}
\end{equation}
\end{small}

\noindent where $f_{v \rightleftharpoons s}\left ( x^{s}_{i} \right )$ maps $x^{s}_{i}$ from visual to semantic FS, then reconstructs it by reversely mapping it from semantic to visual FS. Constraint $f_{v\rightarrow s}\left ( x^{(s)}_{i} \right ) = p_i^{(s)}$ is applied to learn the exact mapping between visual and semantic FS. 

\subsection{Adaptive Adjustment}
Following the cycle mapping, we propose to adaptively adjust the semantic FS in the following steps.

1). For the adjustment of seen class prototypes, we focus on the centroid and current distribution of each instance within the semantic FS:
\begin{small}
\begin{equation}
\begin{aligned}
{p^{(s_{i})}}' = \lambda_{1} p^{(s_{i})} + \gamma_{1} \frac{1}{z}\sum_{j=1}^{z}f_{v\rightarrow s}\left ( x^{(s_{i})}_{j} \right ),
\end{aligned}
\end{equation}
\end{small}

\noindent where ${p^{(s_{i})}}'$ and $p^{(s_{i})}$ are the updated and original prototype of the $i$-th seen class, respectively; $x^{(s_{i})}_{j}$ is the instance and $z$ is the number of instances belonging to this class; $f_{v\rightarrow s}\left ( x^{(s_{i})}_{j} \right )$ calculates the semantic features of $x^{(s_{i})}_{j}$; $\lambda_{1}$, $\gamma_{1}$ are two hyper-parameters that control the balance of these two terms.

2). For the adjustment of unseen class prototypes, because unseen class instances are not available during training in ZSR, we cannot adjust the prototypes straightway. Instead, we propose to adjust the unseen class prototypes by associating with seen class prototypes:
\begin{small}
\begin{equation}
\begin{aligned}
{p^{(u_{i})}}' = \lambda_{2} p^{(u_{i})} + \gamma_{2} \sum_{j=1}^{k} \frac{\Omega (p^{(u_{i})}, p^{(s_{j})})}{{\sum }\Omega}\cdot p^{(s_{j})}, 
\end{aligned}
\end{equation}
\end{small}

\noindent where ${p^{(u_{i})}}'$ and $p^{(u_{i})}$ are the updated and original prototype of the $i$-th unseen class, respectively; $p^{(s_{j})}$ ($j\in \left [ 1, k \right ]$) are the $k$ nearest seen class prototype neighbours of $p^{(u_{i})}$; $\Omega(\cdot,\cdot )$ is a similarity measurement; $\lambda_{2}$, $\gamma_{2}$ are also two hyper-parameters.

3). For the adjustment of global data distribution, we propose a regularization term which considers both the diversity among different class instances and the identity within same class instances:
\begin{small}
\begin{equation}
\begin{aligned}
r = \sum_{i=1}^{n}\sum_{j=1}^{d}\left \| f_{v\rightarrow s}(x^{(s_{i})}_{j}) - O_{i} \right \|^{2},
\end{aligned}
\end{equation}
\end{small}

\noindent where $n$ is the number of seen classes, $d$ is the number of instances belonging to $i$-th seen class $s_{i}$, $O_{i}$ is the semantic centroid of the $i$-th seen class, which can be calculated by $\frac{1}{d} \underset{j} {\sum } f_{v\rightarrow s}(x^{(s_{i})}_{j})$.

\begin{table*}[ht]
    \renewcommand\thetable{2}
    \centering
    \fontsize{8.5}{4}\selectfont 
    \begin{threeparttable}  
        \caption{Comparison with State-of-the-art Competitors}  
        \label{table2}  
        \begin{tabular}{lcccccccccc}  
            \toprule  
            \multirow{2}{*}{Method}&  
            \multicolumn{2}{c}{AWA}&\multicolumn{2}{c}{CUB}&\multicolumn{2}{c}{aPa\&Y}&\multicolumn{2}{c}{ImageNet}\cr  
            \cmidrule(lr){2-3} \cmidrule(lr){4-5} \cmidrule(lr){6-7} \cmidrule(lr){8-9}
            &SS  &ACC               &SS   &ACC              &SS   &ACC                 &SS   &ACC\cr  
            \midrule  
            DeViSE \cite{frome2013devise} ($'13$)               &A/W &56.7/50.4     &A/W  &33.5     &-    &-     &A/W  &12.8\cr
            DAP \cite{lampert2014attribute} ($'14$)             &A   &60.1          &A    &-        &A    &38.2  &-    &-\cr
            MTMDL \cite{yang2014unified} ($'14$)                &A/W &63.7/55.3     &A/W  &32.3     &-    &-     &-    &-\cr
            ESZSL \cite{romera2015embarrassingly} ($'15$)       &A   &75.3          &A    &48.7     &A    &24.3  &-    &-\cr
            SSE \cite{zhang2015zero} ($'15$)                    &A   &76.3          &A    &30.4     &A    &46.2  &-    &-\cr
            RRZSL \cite{shigeto2015ridge} ($'15$)               &A   &80.4          &A    &52.4     &A    &48.8  &W    &-\cr
            Ba et al. \cite{ba2015predicting} ($'15$)           &A/W &69.3/58.7     &A/W  &34.0     &-    &-     &-    &-\cr    
            AMP \cite{bucher2016improving} ($'16$)              &A+W &66.0          &A+W  &-        &-    &-     &A+W  &13.1\cr
            JLSE \cite{zhang2016zero} ($'16$)                   &A   &80.5          &A    &41.8     &A    &50.4  &-    &-\cr
            SynC$^{struct}$ \cite{changpinyo2016synthesized} ($'16$) &A   &72.9     &A    &54.4     &-    &-     &-    &-\cr
            MLZSC \cite{bucher2016improving} ($'16$)            &A   &77.3          &A    &43.3     &-    &53.2  &-    &-\cr
            SS-voc \cite{fu2016semi} ($'16$)                    &A/W &78.3/68.9     &A/W  &-        &-    &-     &A/W  &16.8\cr
            SAE \cite{kodirov2017semantic} ($'17$)              &A   &84.7          &A    &61.2     &A    &55.1  &W    &26.3\cr
            CVAE-ZSL \cite{mishra2017generative} ($'17$)        &A   &71.4          &A    &52.1     &-    &-     &-    &-\cr        
            CLN+KRR \cite{long2017zero} ($'17$)                 &A   &81.0          &A    &58.6     &-    &-     &-    &-\cr
            MFMR \cite{xu2017matrix} ($'17$)                    &A   &76.6          &A    &46.2     &A    &46.4  &-    &-\cr
            RELATION NET \cite{yang2018learning} ($'18$)        &A   &84.5          &A    &62.0     &-    &-     &-    &-\cr
            CAPD-ZSL \cite{rahman2018unified} ($'18$)           &A   &80.8          &A    &45.3     &A    &55.0  &W    &23.6\cr
            {\bf Ours}                                          &A   &{\bf 88.8}    &A    &{\bf 64.7}     &A    &{\bf 56.2}     &W    &{\bf 27.1}\cr  
            \bottomrule  
        \end{tabular}
        \footnotesize{SS: Semantic Space, A: Attributes, W: Word Vectors. '/' and '+' stand for 'or' and 'and' respectively. '-': No comparison/reported results. ACC: Accuracy (\%) where Hit@1 is used for AWA, CUB and aPa\&Y, Hit@5 is used for ImageNet}
    \end{threeparttable}  
\end{table*}  

\subsection{Unified Framework}
Our model optimizes alternately by Eqs. (3)-(6). Specifically, we first optimize Eq. (3) to obtain an initial weight of mapping function. Then Eqs. (4), (5) are performed to adjust prototypes. Lastly, Eqs. (3), (6) are jointly optimized to obtain the updated weight and adaptively adjust the global distribution at the same time. These steps are performed iteratively to reach an optimum (Fig. 1).

The adaptive adjustment can be formulated and combined with cycle mapping, i.e., an encoder-decoder structural model, to a unified framework. Combined with Eqs. (3), (6), the overall objective can be described as:
\begin{small}
\begin{equation}
\begin{aligned}
& J = \sum_{i=1}^{m} \left \| x^{(s)}_{i} - f_{v \rightleftharpoons s}\left ( x^{(s)}_{i} \right ) \right \|^{2} + \alpha\sum_{i=1}^{n}\left \| f_{v\rightarrow s}(x^{(s)}_{i}) - O_{{y^{(s)}_{i}}} \right \|^{2}, \\
& s.t.\quad f_{v\rightarrow s}\left ( x^{(s)}_{i} \right ) = p_i^{(s)}.
\end{aligned}
\end{equation}
\end{small}

We use a hyper-parameter $\alpha$ to balance the importance of these two terms. To simplify, we rewrite the objective function to matrix form:
\begin{small}
\begin{equation}
\begin{aligned}
& J = \left \| \mathbf{X} - \mathbf{W}'\mathbf{W}\mathbf{X}\right \|^2 + \alpha\left \| \mathbf{W}\mathbf{X} - \mathbf{O} \right \|^2,   \\
& s.t.\quad  \mathbf{W}\mathbf{X} = \mathbf{P},
\end{aligned}
\end{equation}
\end{small}

\noindent where $\mathbf{W}$ and $\mathbf{W}'$ are the mapping weights of $f_{v\rightarrow s}(\cdot)$ and $f_{s\rightarrow v}(\cdot)$, respectively. Moreover, to further optimize, we use tied weights \cite{boureau2008sparse} to half the parameters. Then by substituting $\mathbf{W}\mathbf{X}$ with $\mathbf{P}$, our objective can be rewritten as:
\begin{small}
\begin{equation}
\begin{aligned}
&J = \frac{1}{2} \left \| \mathbf{X} - \mathbf{W}^\top \mathbf{P}\right \|^2 + \frac{\alpha}{2} \left \| \mathbf{W}\mathbf{X} - \mathbf{O} \right \|^2,   \\
&s.t.\quad  \mathbf{W}\mathbf{X} = \mathbf{P}.
\end{aligned}
\end{equation}
\end{small}

Eq. (9) is with a hard constraint $\mathbf{W}\mathbf{X} = \mathbf{P}$ that is not easy to solve efficiently. So we relax it to $\beta \left \| \mathbf{W}\mathbf{X} - \mathbf{P} \right \|^2$, where $\beta$ is also a hyper-parameter. We also use trace properties $\mathrm{Tr}(\mathbf{X}) = \mathrm{Tr}(\mathbf{X}^\top)$ and $\mathrm{Tr}(\mathbf{W}^\top \mathbf{P}) = \mathrm{Tr}(\mathbf{P}^\top \mathbf{W})$, then the objective can be further rewritten as:
\begin{small}
\begin{equation}
J =  \frac{1}{2}\left \| \mathbf{X}^\top - \mathbf{P}^\top \mathbf{W}\right \|^2 + \frac{\alpha}{2} \left \| \mathbf{W}\mathbf{X} - \mathbf{O} \right \|^2 + \frac{\beta}{2} \left \| \mathbf{W}\mathbf{X} - \mathbf{P} \right \|^2.
\end{equation}
\end{small}

To solve it, we take a derivative of Eq. (10) with respect to $\mathbf{W}$, and set it to zero, i.e., 
\begin{small}
\begin{equation}
\begin{aligned}
\mathbf{P}\mathbf{P}^\top \mathbf{W} + (\alpha + \beta)\mathbf{W}\mathbf{X}\mathbf{X}^\top - \left[(1+\beta)\mathbf{P} + \alpha \mathbf{O} \right ]\mathbf{X}^\top = \mathbf{0}.
\end{aligned}
\end{equation}
\end{small}

We denote $\mathbf{L} = \mathbf{P}\mathbf{P}^\top$, $\mathbf{R} = (\alpha + \beta)\mathbf{X}\mathbf{X}^\top$ and $\mathbf{M} = - \left[(1+\beta)\mathbf{P} + \alpha \mathbf{O} \right ]\mathbf{X}^\top$. Therefore, Eq. (11) can be rewritten as:
\begin{small}
\begin{equation}
\mathbf{L}\mathbf{W} + \mathbf{W}\mathbf{R} + \mathbf{M} = \mathbf{0}, 	
\end{equation}
\end{small}

\noindent which is exactly in the standard form of the generalized Lyapunov equation and can be solved efficiently by an existing solver \cite{pomet1992adaptive}.

\begin{figure*}[ht]  
	\begin{minipage}[t]{0.33\linewidth}  
		\centering  
		\includegraphics[width=2.2in]{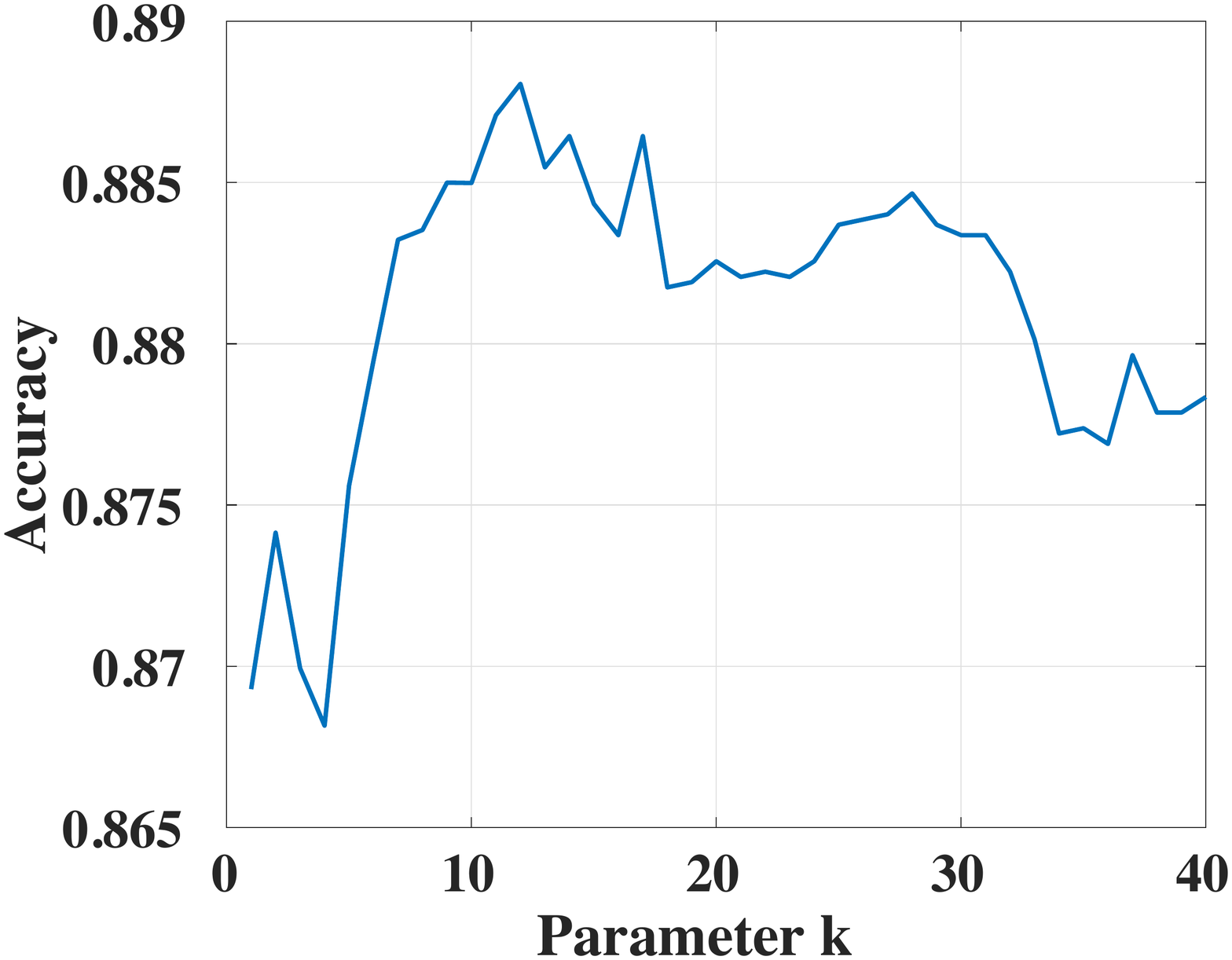}  
		\caption{K Search for AWA}  
		\label{fig3}  
	\end{minipage}
	\begin{minipage}[t]{0.33\linewidth}  
		\centering  
		\includegraphics[width=2.2in]{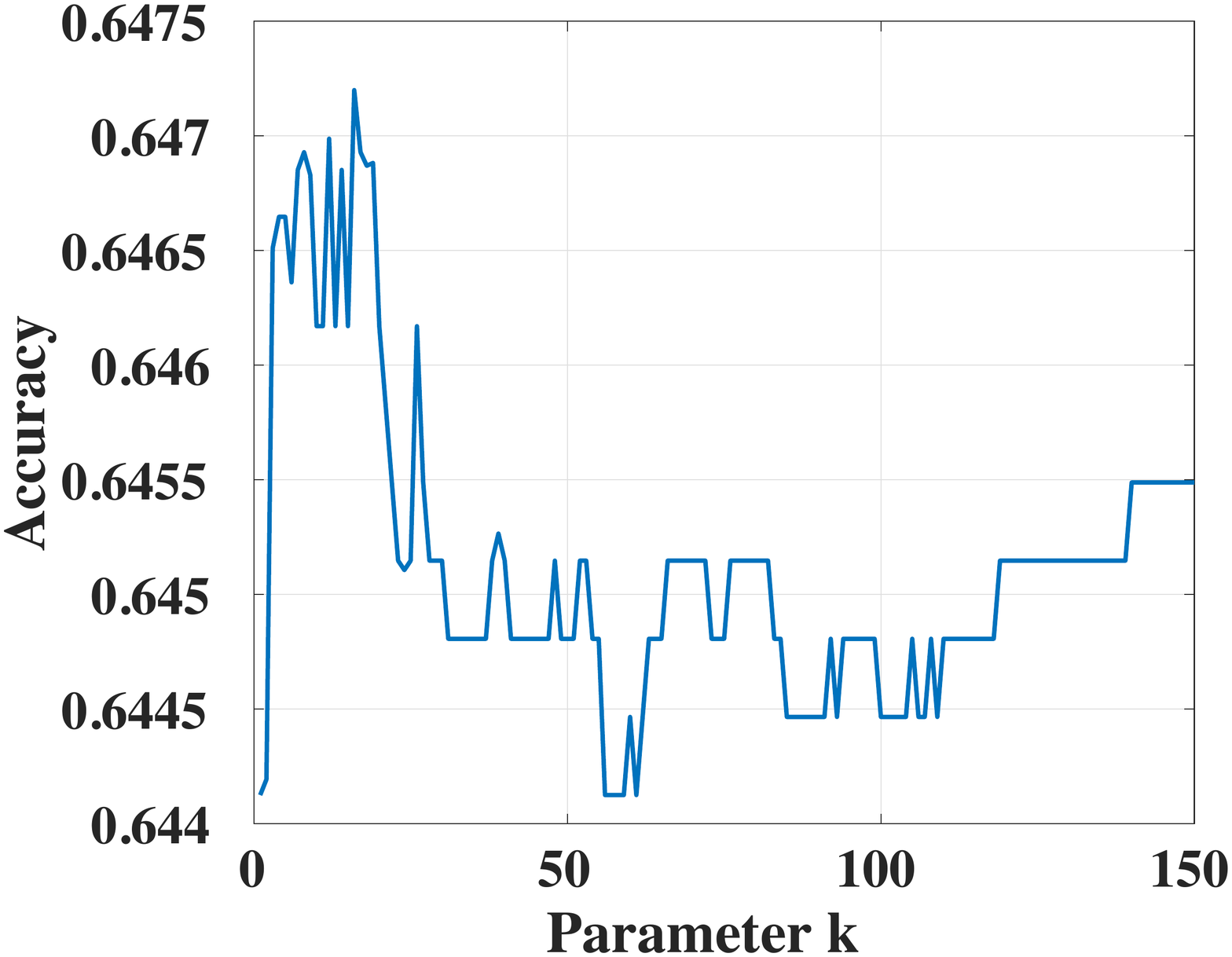} 
		\caption{K Search for CUB}  
		\label{fig4}  
	\end{minipage} 
	\begin{minipage}[t]{0.33\linewidth}  
		\centering  
		\includegraphics[width=2.2in]{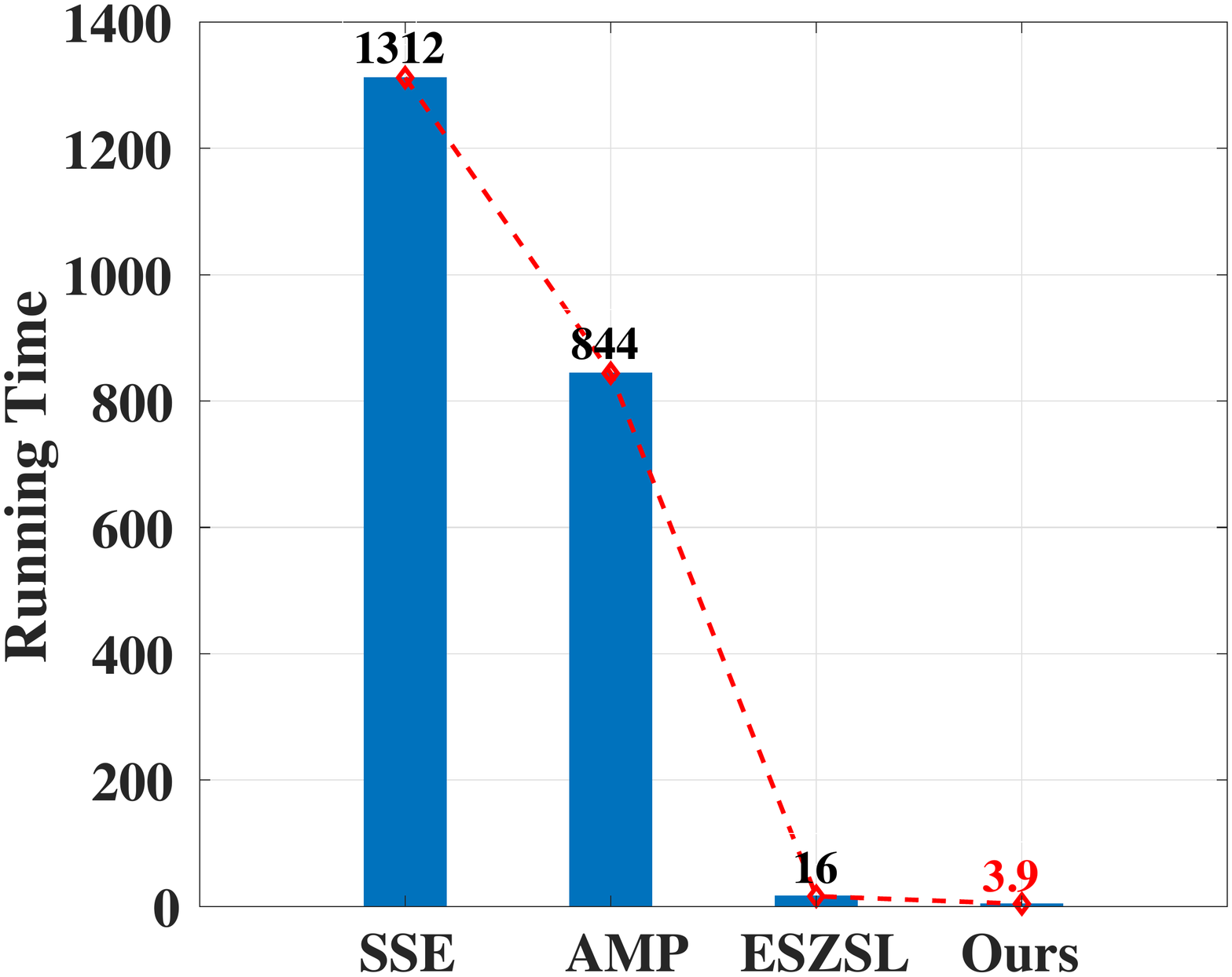}
		\caption{Trainning Time}  
		\label{fig5}  
	\end{minipage}   
\end{figure*}

\section{Experiment}
\subsection{Dataset and Setting}
Our model is evaluated on four benchmark datasets in ZSR, including Animals with Attributes (AWA) \cite{lampert2014attribute}, CUB-200-2011 Birds (CUB) \cite{wah2011caltech}, aPascal\&Yahoo (aPa\&Y) \cite{farhadi2009describing} and ILSVRC2012/ILSVRC2010 (ImageNet) \cite{russakovsky2015imagenet}. We adopt Hit@k accuracy \cite{frome2013devise} to evaluate the performance. It is a widely used evaluation criterion in ZSR, and it refers to predict top-k possible class labels of the testing unseen class instance. The model classifies the instance correct, if and only if the ground truth is within these top-k class labels. Following most methods, the datasets \& settings are shown in Table 1. 
\begin{table}[htbp]
\renewcommand\thetable{1}
\begin{center}
\caption{Dataset Settings. SC/UC: Seen/Unseen Class; SS: Semantic Space, A: Attributes, W: Word Vectors.}
		\setlength{\tabcolsep}{0.8mm}{ 
\begin{tabular}{|c|c|c|c|c|c|}
\hline
Dataset  & Instances         & SC   & UC  & SS & Accuracy  \\ \hline
AWA      & 30475             & 40   & 10  & A & Hit@1   \\ \hline
CUB      & 11788             & 150  & 50  & A & Hit@1   \\ \hline
aPa\&Y   & 15339             & 20   & 12  & A & Hit@1   \\ \hline
ImageNet & $2.54\times 10^5$ & 1000 & 360 & W & Hit@5   \\ \hline
\end{tabular}
}
\end{center}
\end{table}

The features are extracted from GoogleNet \cite{szegedy2015going} for the visual feature space. Each image instance is presented by a 1024-dimensional vector. In our model, the cosine similarity is adopted for $\Omega(\cdot,\cdot )$. Hyper-parameter $\lambda_{1}$/$\gamma_{1}$ are set to 0.75/0.25, and $\lambda_{2}$/$\gamma_{2}$ are set to 0.8/0.2, respectively by grid-search \cite{hsu2003practical}. Our model strictly complies with the zero-shot setting that the training of mapping function only relies on seen classes. All selected comparison methods are under the same settings.
\subsection{Results and Analysis}
The comparison results are shown in Table 2. We can see our model outperforms all competitors with great advantages. The accuracy achieves 88.8\%, 64.7\%, 56.2\% and 27.1\% for AWA, CUB, aPa\&Y and ImageNet respectively. Next, we introduce some further analysis. Firstly, we explore the influence of parameter k, which refers to the k nearest neighbors used in the adjustment for unseen class prototypes (Eq. (5)). We evaluate on AWA and CUB for a smaller and larger k-search respectively, and the results are shown in Fig.2 and Fig.3. For AWA, the preferred range is $k\in [7,17]$ and the most preferred $k$ is around 12. For CUB, the preferred range is $k\in [4,19]$ and the most preferred $k$ is around 16. Secondly, we evaluate the training time. We compare our model with ESZSL \cite{romera2015embarrassingly}, SSE \cite{zhang2015zero} and AMP \cite{bucher2016improving}. The results are shown in Fig.4. We can see that the training of our model is much faster than these competitors. 

\section{Conclusion}
In this paper, we propose a novel model based on a unified learning framework for zero-shot recognition. Our model adaptively adjusts semantic feature space by considering both the prototypes and the global distribution of data. Moreover, our model can be formulated to a more efficient framework and significantly boosts the training. Extensive experiments verified the effectiveness of our model and obtained remarkable performance compared with other existing representative methods.

\bibliographystyle{IEEEbib}
\bibliography{my}
\end{document}